\documentclass{article}

\usepackage[final]{nips_2018}

\usepackage[utf8]{inputenc} 
\usepackage[T1]{fontenc}    
\usepackage{hyperref}       
\usepackage{url}            
\usepackage{booktabs}       
\usepackage{amsfonts}       
\usepackage{amsmath}       
\usepackage{nicefrac}       
\usepackage{microtype}      
\usepackage{graphicx}

\title{Copy the Old or Paint Anew? An Adversarial Framework for (non-) Parametric Image Stylization}
\author{
  Nikolay Jetchev\\
  \footnotesize{Zalando Research}\\
 \footnotesize{ Berlin, Germany} \\
  \footnotesize{\texttt{nikolay.jetchev@zalando.de}}\\
  \And
  Urs Bergmann\\
   \footnotesize{Zalando Research}\\
   \footnotesize{Berlin, Germany}  \\
   \footnotesize{\texttt{urs.bergmann@zalando.de}} \\
  \And
  Gokhan Yildirim\\
   \footnotesize{Zalando Research}\\
   \footnotesize{Berlin, Germany}  \\
   \footnotesize{\texttt{gokhan.yildirim@zalando.de}} \\
}

\begin{document}

\maketitle
\vspace*{-0.5cm}
\begin{abstract}
Parametric generative deep models are state-of-the-art for photo and non-photo realistic image stylization. However, learning complicated image representations requires 
compute-intense models parametrized by a huge number of weights, which in turn requires large datasets to make learning successful.
Non-parametric exemplar-based generation is a technique that works well to reproduce style from small datasets, but is also compute-intensive. These aspects  are a drawback for the practice of digital AI artists: typically one wants to use a small set of stylization images, and needs a fast flexible model in order to experiment with it.
With this motivation, our work has these contributions:
(i) a novel stylization method called Fully Adversarial Mosaics (FAMOS) that combines the strengths of both parametric and non-parametric approaches;
(ii) multiple ablations and image examples that analyze the method and show its capabilities;
(iii) \href{https://github.com/zalandoresearch/famos}{source code} that will empower artists and machine learning researchers to use and modify FAMOS.

\end{abstract}
\vspace*{-0.3cm}
Tiling of small stones was a classical ancient art form, and in modern times there are efficient algorithms to produce such mosaics (with non-overlapping tiles) digitally \cite{JIM}.
Seamless mosaics in the style of the Renaissance painter Archimboldo are more challenging, but modern deep learning methods allow efficient seamless image stylization. Neural style transfer \cite{GatysEB15a} uses filter statistics (pretrained on a huge dataset) of a style image to optimize an output image. \cite{GANosaic} learns a texture process adversarially from multiple input images and then optimizes samples from it to paint a mosaic.
\cite{pix2pix} can stylize images in one forward pass of the generator network, but requires for training paired images from 2 domains. \cite{cycle} is more flexible and can translate unpaired images. However, the authors note that only local stylization of color and texture is learned, larger geometric changes are not easily achieved. In general, parametric approaches to image generation have capacity issues when very complicated image styles are considered.

Non-parametric image quilting \cite{EfrosQ} combines patches from texture data in order to smoothly reconstruct a target image -- "texture transfer".
The work of \cite{analogy17} uses patch similarity and content copying in order to solve an image analogy problem.
The work of \cite{liwand16} uses neural patch matching (hard attention) combined with a content loss. Their results are visually impressive, but expensive, and for one output image only.
Both \cite{liwand16,analogy17} rely on pretrained networks and this can negatively impact the image quality if the texture and content image distributions differ too much.
As a further drawback, all of \cite{EfrosQ,liwand16,analogy17} are slow due to nearest neighbours routines used for patch search, and cannot scale to large images.

There are already methods to improve conditional parametric image generation with a non-parametric memory \cite{pixelnn,sims}, but they are more specific for datasets of paired images, and are not directly suitable for artistic applications of image stylization. These methods are also slow to optimize a single output image -- the bottleneck is the need to perform expensive search in the database.



\begin{figure}[t]
 \setlength{\abovecaptionskip}{0pt}
\setlength{\belowcaptionskip}{0pt} 
    \centering
    \includegraphics[width=6.5cm]{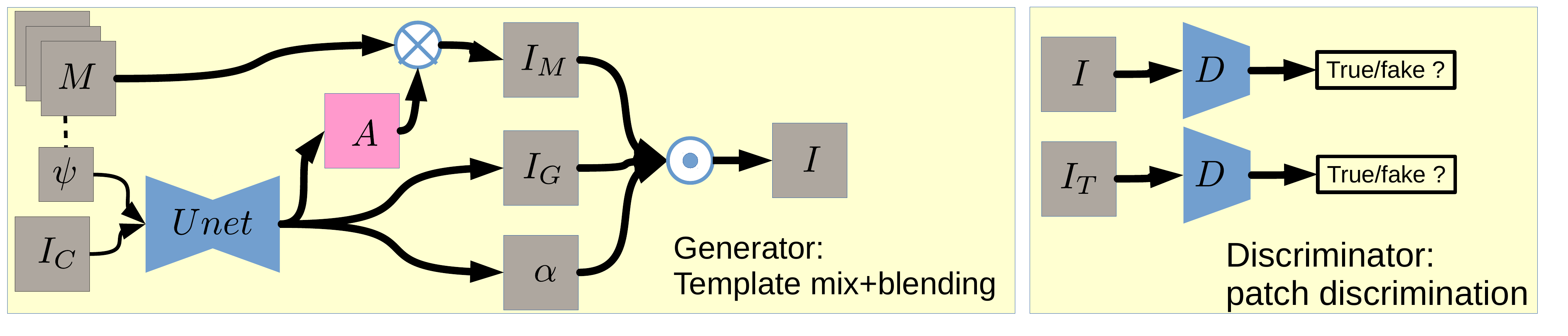}
    \caption{\footnotesize{The architecture of the generator $G$: a single Unet predicting the template mixture weights $A$, an image $I_G$ and blending mask $\alpha$. The discriminator predicts true/fake looking texture patches.} }
    \label{fig:arch}
\end{figure}

Our method, called FAMOS, combines the advantages of parametric methods (fast to infer stylization once trained on data) with those of non-parametric methods (accurate reconstruction of complex styles) in a GAN \cite{Goodfellow14} framework.
We learn a fully convolutional network that can predict stylization images of very large size. In Figure \ref{fig:arch} we show the architecture of the generator $G$, which is the key novelty. A U-net \cite{pix2pix} is used to predict a mixture matrix $A$ used to create image $I_M$ copied from input style data, a mask $\alpha$ and an image $I_G$. They are used to blend and fine tune the final output $I= \alpha \odot I_G + (1-\alpha) \odot I_M$. This structure allows our network to adaptively fill some regions of the final image with generated content similar to a traditional methods (e.g. \cite{pix2pix}) and for other regions decide to directly copy from the source style data.

\begin{figure}[t]
 \setlength{\abovecaptionskip}{0pt}
\setlength{\belowcaptionskip}{0pt} 

    \centering
    \includegraphics[height=1.8cm]{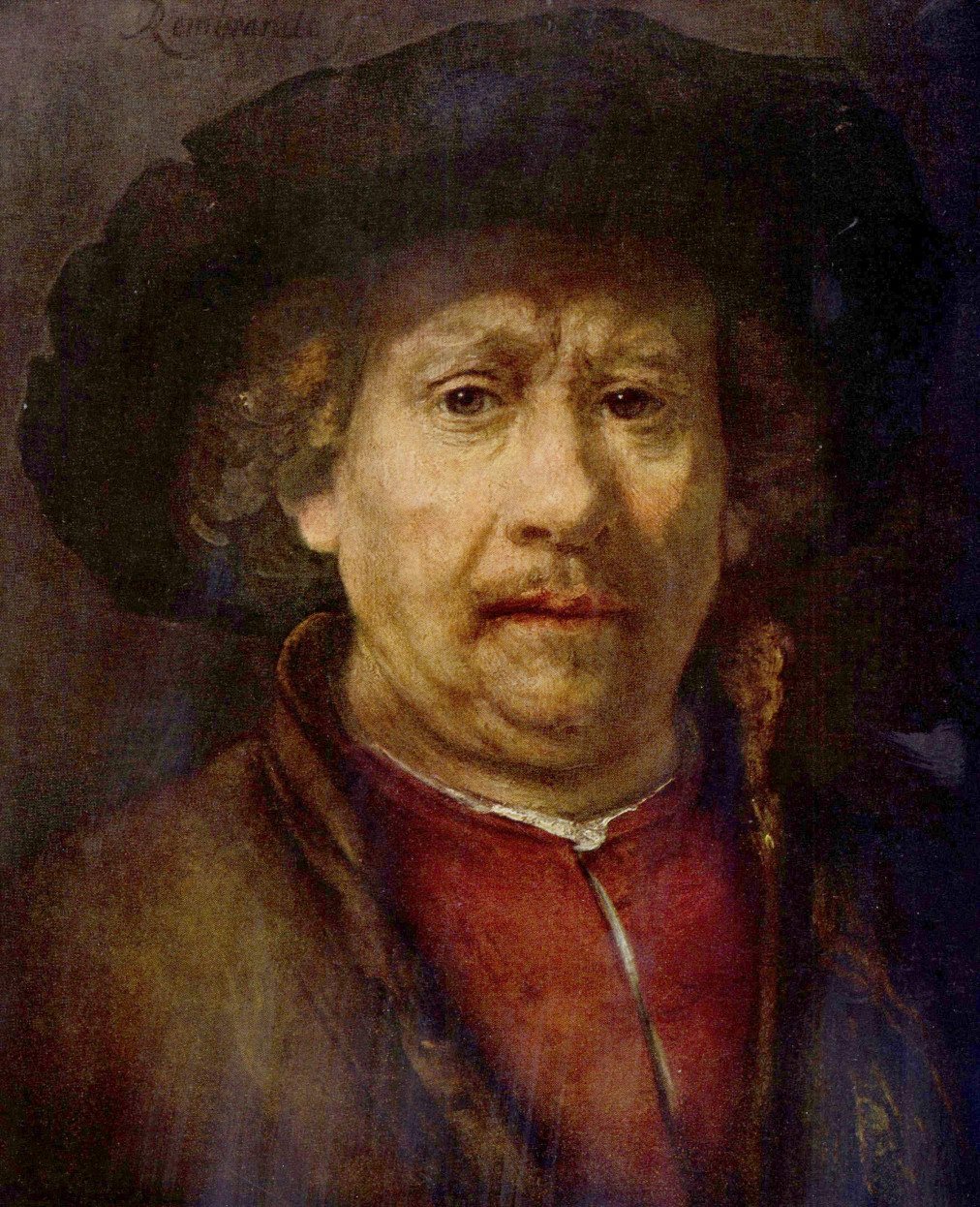}
    \includegraphics[height=3.4cm]{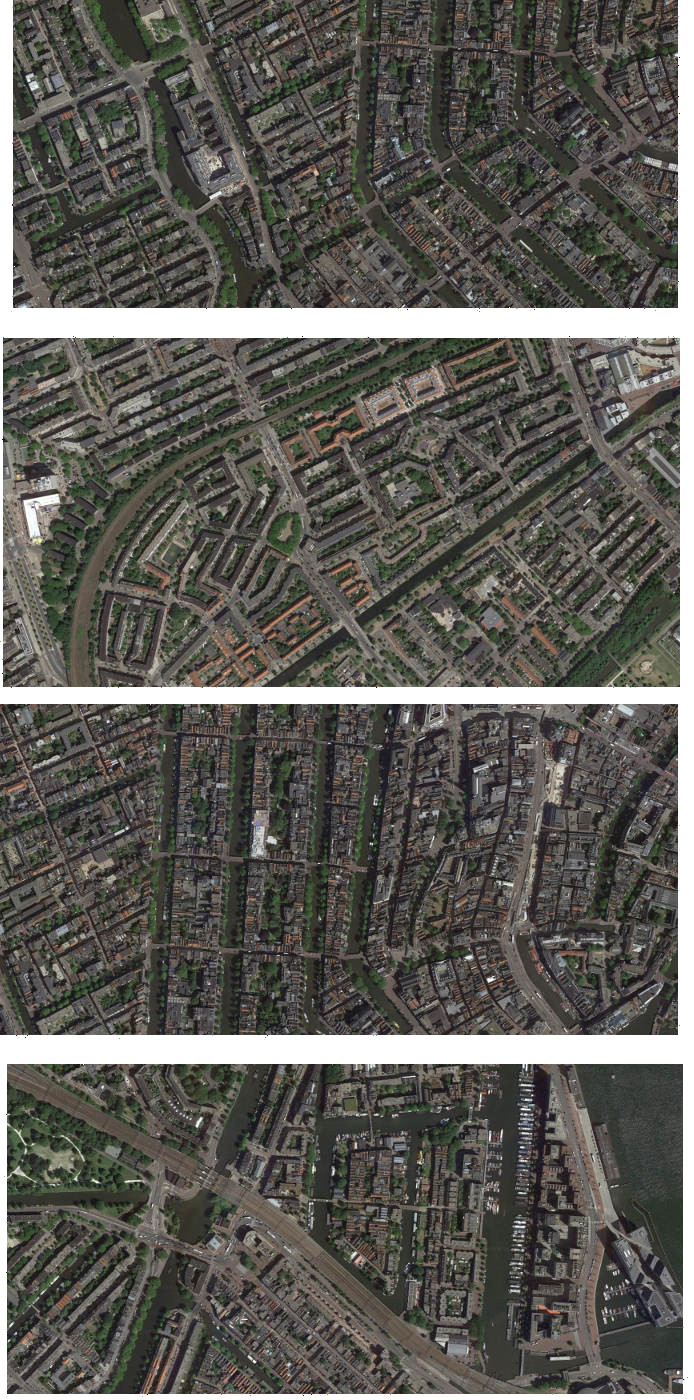}
    \includegraphics[height=3.4cm]{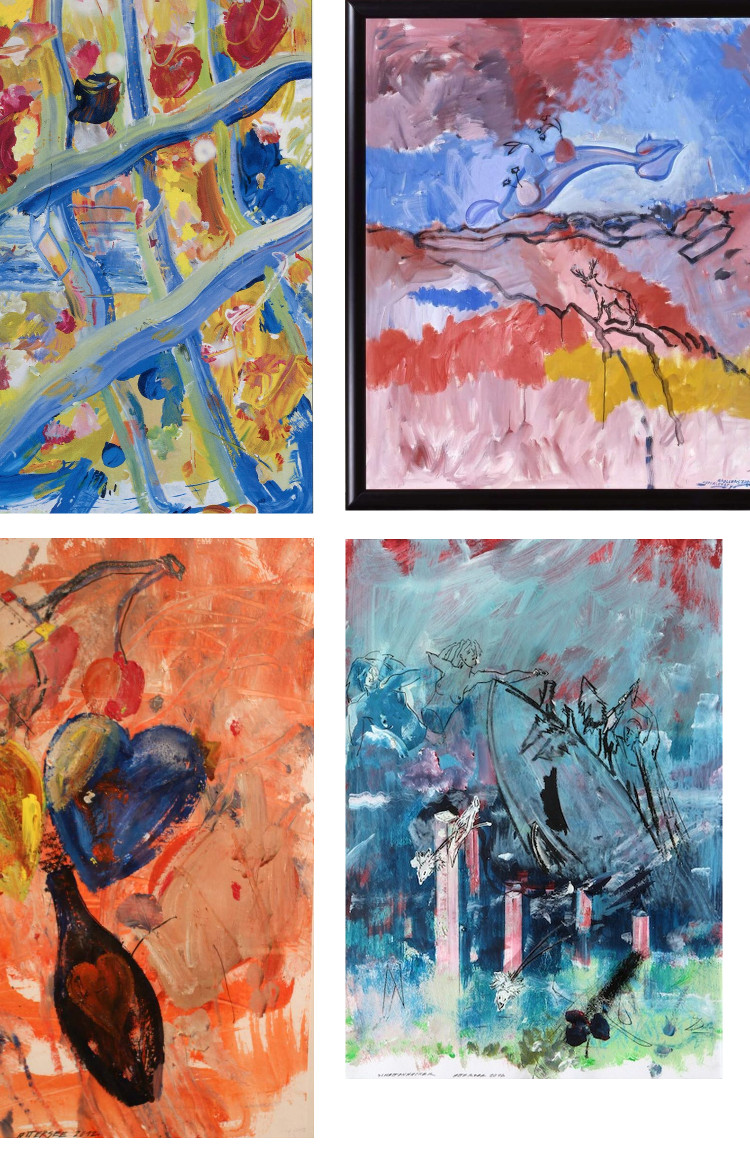}
    \includegraphics[height=3.4cm]{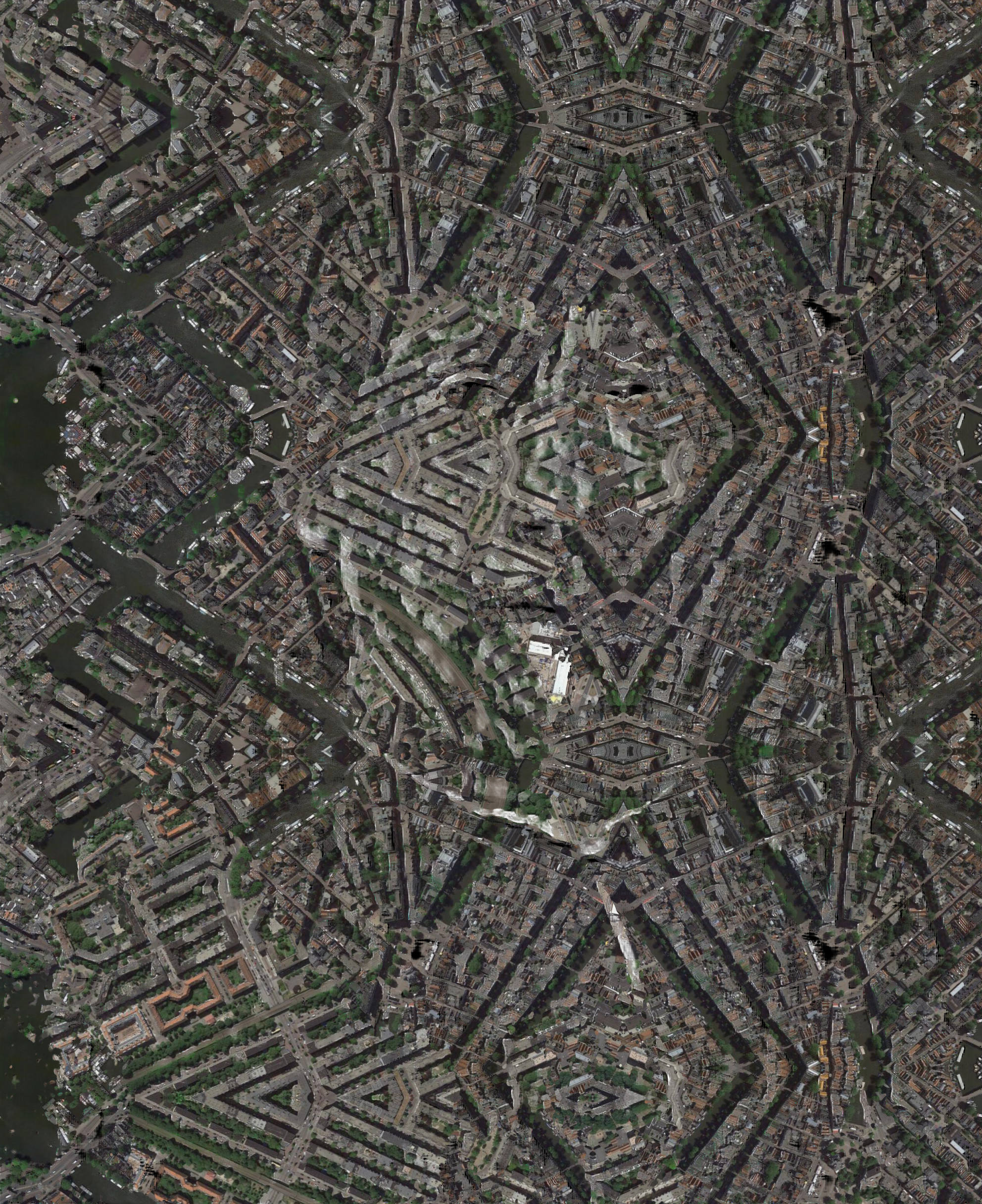}
    \includegraphics[height=3.4cm]{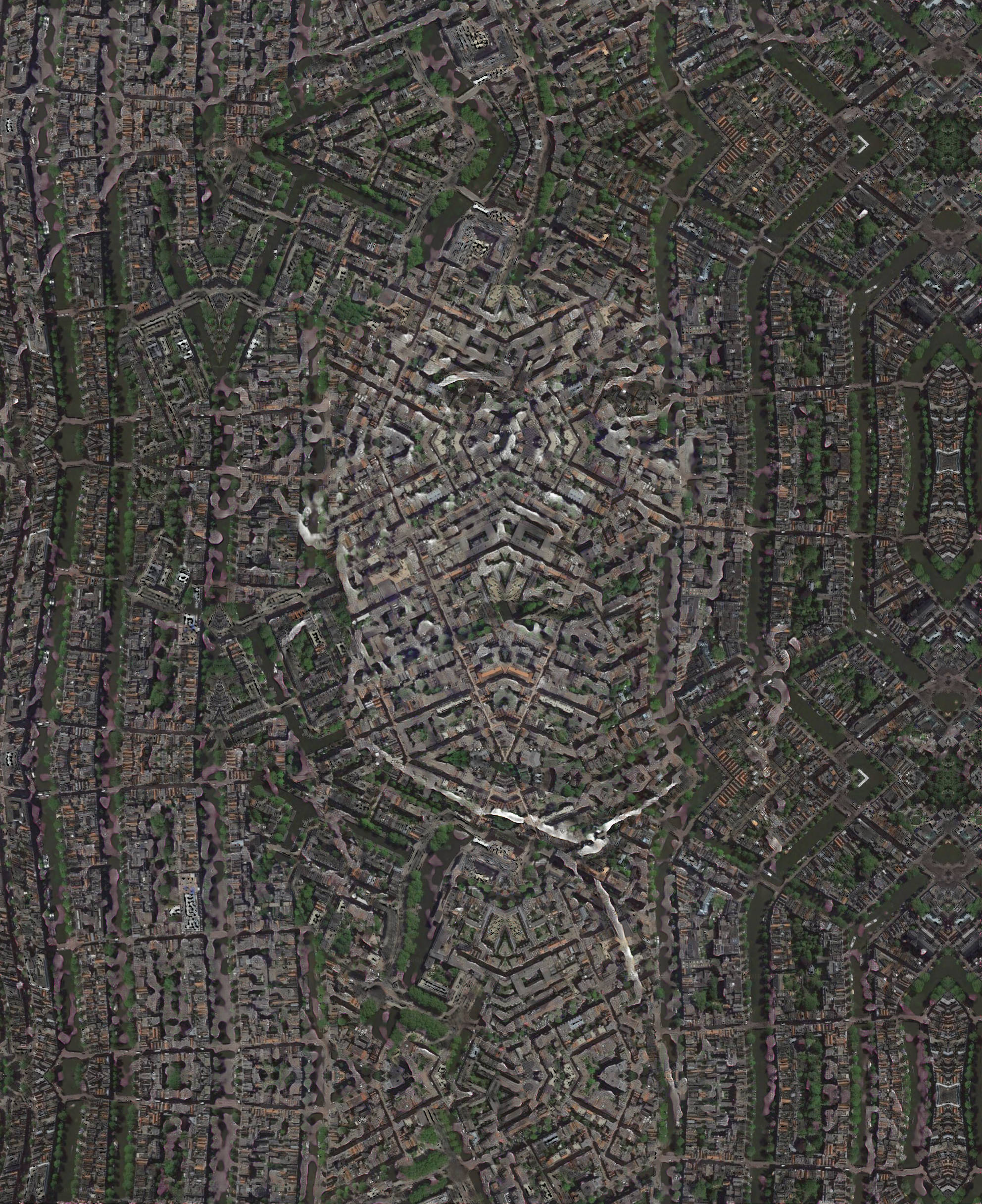}
    \includegraphics[height=3.4cm]{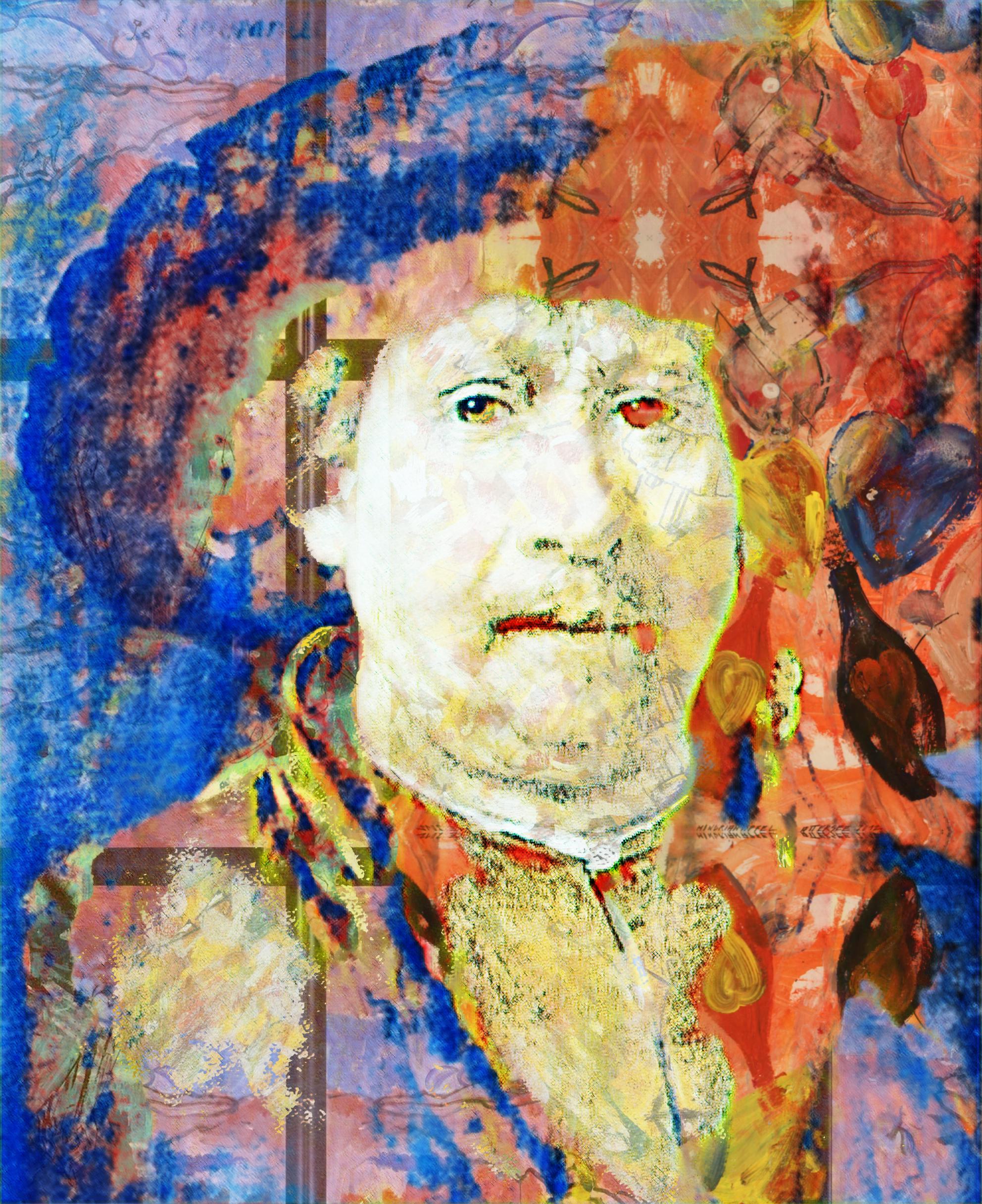}
    \caption{\footnotesize{FAMOS outputs a stylization of a Rembrandt portrait as content image, 2000x2400 pixels. Style images: 8 Google maps views of Amsterdam, and 7 paintings of C. Attersee (4 shown). We can emphasize more or less the texture aspect or content reproduction by tweaking the loss hyperparameter ($\lambda=3$ and $\lambda=1$ shown for Amsterdam).} }
    \label{fig:mosaic}
\end{figure}

For training FAMOS we have as input for the discriminator: (i) a set of \textit{content} images $I_C$ of various sizes, from which we can define the cropped patch distribution $I_c \sim P_C$ of fixed size $H \times W$; (ii) \textit{style} (texture) images $I_T$ with crop distribution $I_t \sim P_T$. For the generator, we crop from the memory $M$ (see below) patches $(M',\psi) \sim P_M$. Here we keep also the crop grid $\psi$, of size $2 \times H \times W$. For training FAMOS, we will use minibatches of patches, and for inference we can roll-out on any image size (due to the fully convolutional generator). The loss function combines adversarial and content reconstruction terms:
 \setlength{\abovedisplayskip}{4pt}
 \setlength{\belowdisplayskip}{4pt}
\begin{align}
\mathcal{L}_{adv}(G,D)&=E_{I_t \sim P_T}\log D(x)+E_{I_c \sim P_C, (M',\psi) \sim P_M}\log(1-D(G(I_c,M',\psi))),\\
\mathcal{L}_{content} &= E_{I_c \sim P_C, M' \sim P_M}\|\phi(I_c)-\phi(G(I_c,M'))  \|_2,\\
G^* &= \arg\min_G\arg\max_D \mathcal{L}_{adv} + \lambda \mathcal{L}_{content}.
\end{align}
The \textit{correspondence map} $\phi$ in $\mathcal{L}_{content}$ specifies the perceptual space for the reconstruction term \cite{EfrosQ,GANosaic}.



 The use of a memory $M\in \mathrm{R}^{N\times3\times H \times W}$ is a novel \textbf{non-parametric aspect} of our model. It is created by randomly sampling $N$ images from $I_T$ and  interpolating each of them into a tiling of a fixed size. Each tiling has a random initial grid coordinate offset, and we may also call them \textit{templates}.
The mixture matrix $A\in \mathrm{R}^{N \times H \times W}  $ is calculated as a function of the content patch $I_c$ and the memory template coordinates $\psi$. In a sense, $\psi$ encodes the available template patches (given the offset) with which we can paint at that spatial position of the convolutional input. Since we use a set of tiled templates $M$, this gives enough information to the network to predict a good mixture of coefficients to reconstruct the content and fool the discriminator.

We apply softmax to $A$ (in the dimension of the $N$ templates) to get $\tilde{A}$, and use it to create the soft attention memory image $I_M$ via batched dot operation $\tilde{A} \otimes M = I_M  \in \mathrm{R}^{3 \times H \times W}$.
While somewhat analogous to various attention-like methods \cite{NonLocal2018}, it is applied only to fixed spatial positions -- e.g. spatial position $h,w$ in the output $I_M$ results from copying values from the same position $h,w$ from $M$. This aligned spatial structure is a cost efficient alternative to full attention or to nearest neighbour neural patch best match search, $O(HWN)$ vs $O(H^2W^2)$. Note that each tiling is translated by a random offset, so if $N \to\infty$ our module will be expressive as full spatial attention. However, in practice a small $N$ is enough to construct an expressive yet not too memory hungry memory module.






In \textbf{summary}, FAMOS is an image stylization approach with a new concept, with the following key properties: (i) generation of \textbf{seamless mosaics} with unique visual aesthetic, a single neural model that can decide whether to copy or generate parametrically different regions of the output image.
(ii) a \textbf{computationally efficient} non-parametric memory module that allows to copy complex textures that are difficult to represent in a parametric model; 
(iii) the fully convolutional model can create \textbf{very large images} at inference time -- all calls to $G$ can be efficiently split into small chunks seamlessly forming a whole image, without memory constraints \cite{GANosaic,SGAN2016}; 
(iv) a  \textbf{flexible framework} that can fit many artistic choices given style and content input data.
The result of training FAMOS and inferring a final mosaic on a large content image is shown in Figure \ref{fig:mosaic} -- zooming-in reveals how well style image details are preserved. See the Appendices for more generative art.




\pagebreak
 \clearpage
 
\bibliography{bibi}

\begin{thebibliography}{10}

\bibitem{pixelnn}
Aayush Bansal, Yaser Sheikh, and Deva Ramanan.
\newblock Pixelnn: Example-based image synthesis.
\newblock {\em CoRR}, abs/1708.05349, 2017.

\bibitem{PSGAN2017}
Urs Bergmann, Nikolay Jetchev, and Roland Vollgraf.
\newblock Learning texture manifolds with the periodic spatial {GAN}.
\newblock In {\em Proceedings of The 34th International Conference on Machine
  Learning}, 2017.

\bibitem{EfrosQ}
Alexei~A. Efros and William~T. Freeman.
\newblock Image quilting for texture synthesis and transfer.
\newblock In {\em Proceedings of the 28th Annual Conference on Computer
  Graphics and Interactive Techniques}, SIGGRAPH, 2001.

\bibitem{GatysEB15a}
Leon~A. Gatys, Alexander~S. Ecker, and Matthias Bethge.
\newblock A neural algorithm of artistic style.
\newblock {\em CoRR}, abs/1508.06576, 2015.

\bibitem{Goodfellow14}
Ian~J. Goodfellow, Jean Pouget{-}Abadie, Mehdi Mirza, Bing Xu, David
  Warde{-}Farley, Sherjil Ozair, Aaron~C. Courville, and Yoshua Bengio.
\newblock Generative adversarial nets.
\newblock In {\em Advances in Neural Information Processing Systems 27}, 2014.

\bibitem{wgan}
Ishaan Gulrajani, Faruk Ahmed, Mart{\'{\i}}n Arjovsky, Vincent Dumoulin, and
  Aaron~C. Courville.
\newblock Improved training of wasserstein gans.
\newblock {\em CoRR}, 2017.

\bibitem{pix2pix}
Phillip Isola, Jun-Yan Zhu, Tinghui Zhou, and Alexei~A Efros.
\newblock Image-to-image translation with conditional adversarial networks.
\newblock {\em CoRR}, abs/1611.07004, 2016.

\bibitem{GANosaic}
Nikolay Jetchev, Urs Bergmann, and Calvin Seward.
\newblock Ganosaic: Mosaic creation with generative texture manifolds.
\newblock {\em CoRR}, abs/1712.00269, 2017.

\bibitem{SGAN2016}
Nikolay Jetchev, Urs Bergmann, and Roland Vollgraf.
\newblock Texture synthesis with spatial generative adversarial networks.
\newblock {\em CoRR}, abs/1611.08207, 2016.

\bibitem{JIM}
J.~Kim and F.~Pellacini.
\newblock Jigsaw image mosaics.
\newblock In {\em Proc. of the 29th Annual Conference on Computer Graphics and
  Interactive Techniques}, SIGGRAPH, 2002.

\bibitem{KingmaB14}
Diederik~P. Kingma and Jimmy Ba.
\newblock Adam: A method for stochastic optimization.
\newblock {\em CoRR}, abs/1412.6980, 2014.

\bibitem{liwand16}
Chuan Li and Michael Wand.
\newblock Combining markov random fields and convolutional neural networks for
  image synthesis.
\newblock In {\em {CVPR}}, pages 2479--2486. {IEEE} Computer Society, 2016.

\bibitem{analogy17}
Jing Liao, Yuan Yao, Lu~Yuan, Gang Hua, and Sing~Bing Kang.
\newblock Visual attribute transfer through deep image analogy.
\newblock {\em {ACM} Trans. Graph.}, 36(4):120:1--120:15, 2017.

\bibitem{RadfordMC15}
A.~Radford, L.~Metz, and S.~Chintala.
\newblock Unsupervised representation learning with deep convolutional
  generative adversarial networks.
\newblock {\em CoRR}, abs/1511.06434, 2015.

\bibitem{NonLocal2018}
Xiaolong Wang, Ross Girshick, Abhinav Gupta, and Kaiming He.
\newblock Non-local neural networks.
\newblock {\em CVPR}, 2018.

\bibitem{sims}
Jiaya~Jia Xiaojuan~Qi, Qifeng~Chen and Vladlen Koltun.
\newblock Semi-parametric image synthesis.
\newblock 2018.

\bibitem{cycle}
Jun-Yan Zhu, Taesung Park, Phillip Isola, and Alexei~A Efros.
\newblock Unpaired image-to-image translation using cycle-consistent
  adversarial networks.
\newblock In {\em Computer Vision (ICCV), 2017 IEEE International Conference
  on}, 2017.

\bibitem{modal}
Jun-Yan Zhu, Richard Zhang, Deepak Pathak, Trevor Darrell, Alexei~A Efros,
  Oliver Wang, and Eli Shechtman.
\newblock Toward multimodal image-to-image translation.
\newblock In {\em Advances in Neural Information Processing Systems 30}, 2017.

\end{thebibliography}
\setcitestyle{numbers}
\bibliographystyle{plain}

\pagebreak
 \clearpage
\section*{Appendix I: Exploration of FAMOS Possibilities}
\label{app:abbla}

\begin{figure}[t]
    \centering
    \includegraphics[height=5.cm]{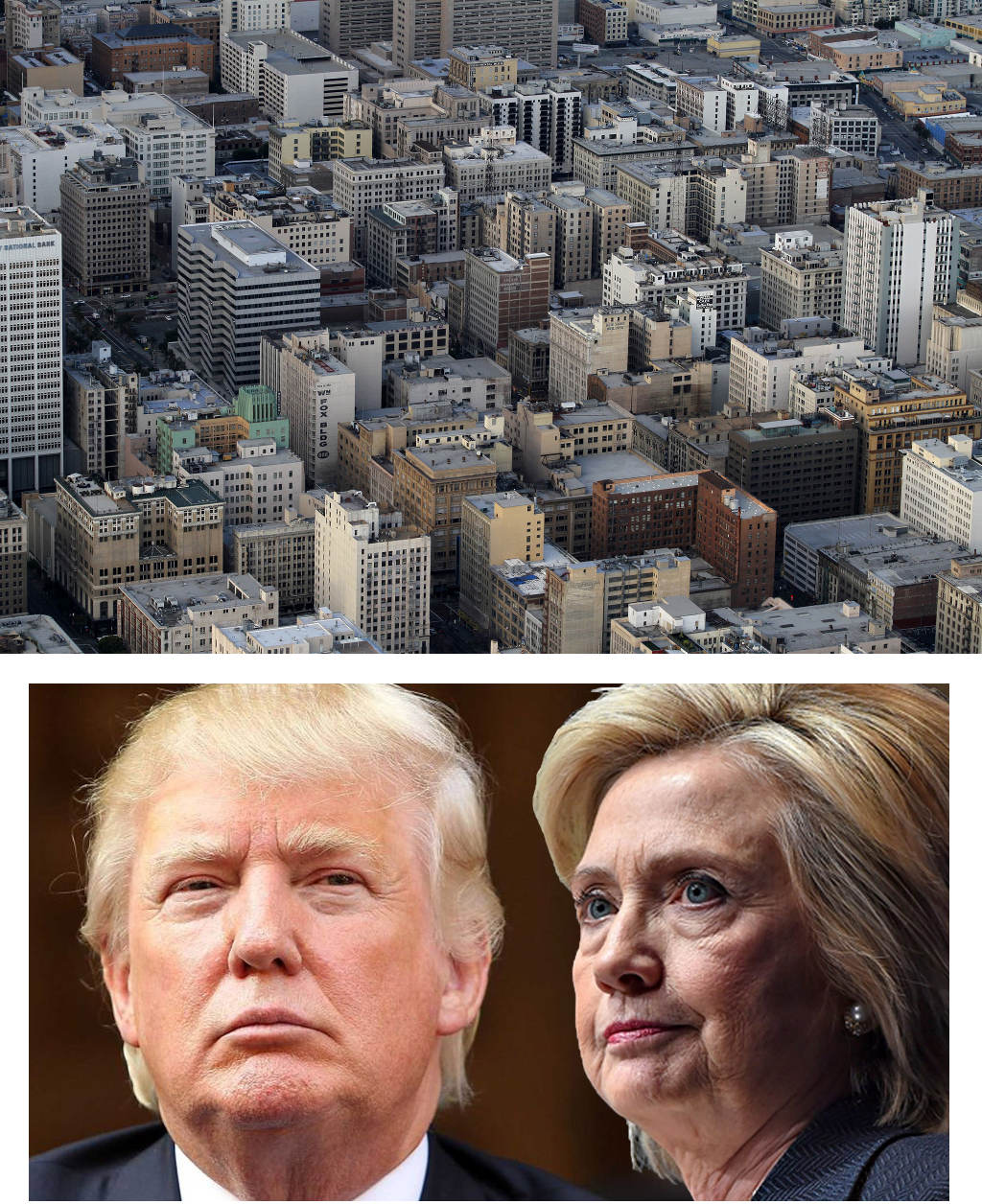}
    \includegraphics[height=6cm]{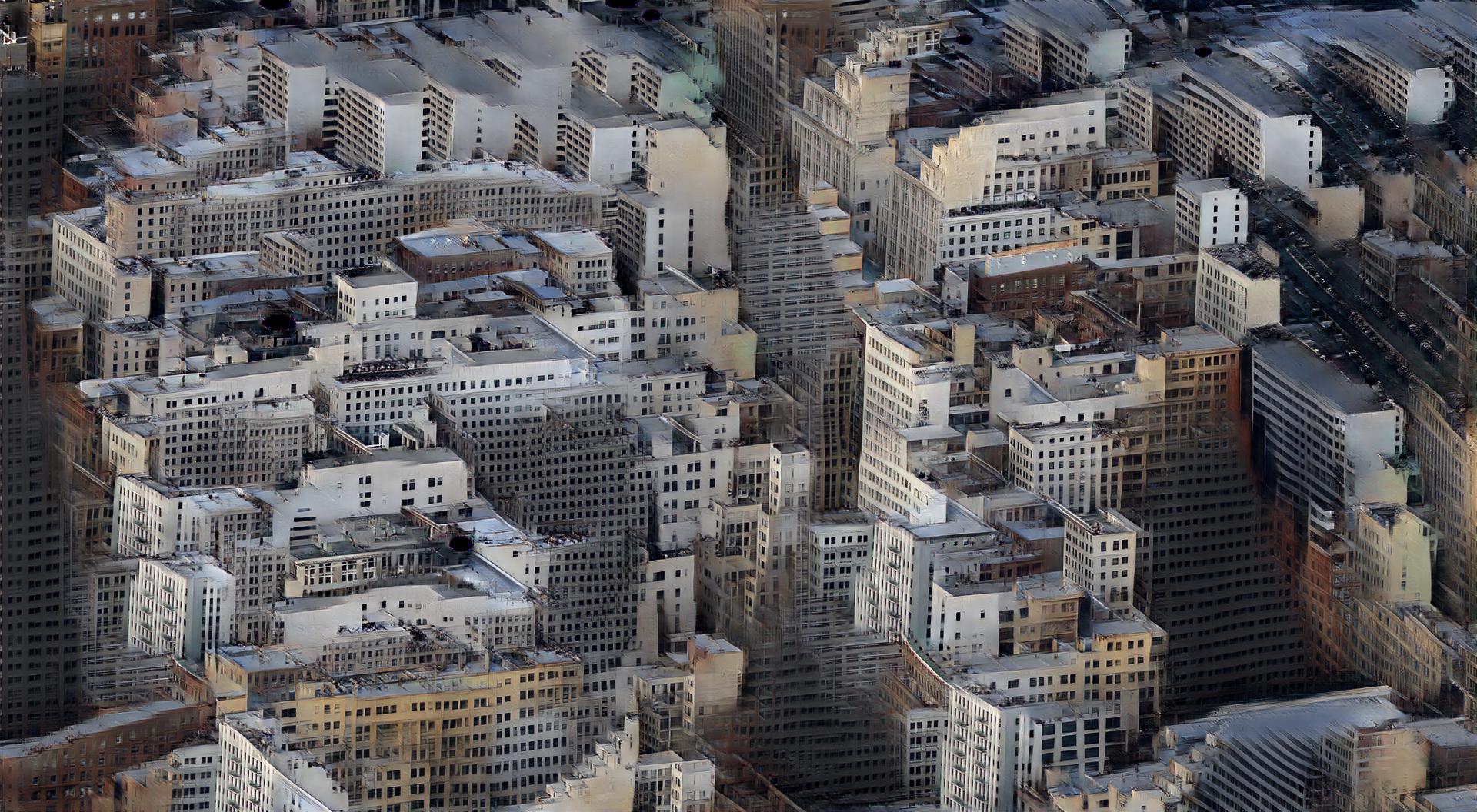}
    \caption{A content image of certain two famous people stylized with an aerial view of downtown Los Angeles. Mosaic image size 1920x1080 pixels. Image created with a FAMOS model without memory templates -- the regular nature of the city grid allowed nice results with the convolutional Unet generator.}
    \label{fig:la}
\end{figure}

\paragraph{Tweaking FAMOS: parametric vs non-parametric modules} 

Our tool can deliver good looking mosaics, but it is still a tool dependent on experimentation done by the artist. The \textit{right} choice of texture and content images is a part of the creative search for a good artwork. The selection of the neural architecture and the hyperparameters can also change the final result a lot. Here we give a small summary of all the choices the user of FAMOS can do in order to influence the final artwork:

\begin{itemize}
\item choice of content and texture image sets, potentially scaling them to tweak scale of visual details relative to generator receptive field (see \cite{SGAN2016} for examples). We get best results when using few  style images with consistent texture properties (repetition of small details), but large sets of diverse images may also work.
\item the most important choice: keep only the parametric or non-parametric generative modules of the network, discussed in detail in the current section.
\item layer count and kernel size of the $G$ and $D$ networks, changing the receptive field of the network
\item Correspondence map for image reconstruction distance metric -- see below
\item count of memory templates $N$ and choice of tiling mode (see below)
\item regularization of mask $\alpha$, affecting how much the FAMOS relies on copying non-parametrically and parametrically. See Appendix II for more on this and other regularization terms we considered.
\item stochastic noise -- we add also noise to the bottleneck of the Unet, see \cite{PSGAN2017} for some intuition how this can change the generated output.
\end{itemize} 

We share the links for two online galleries we prepared with many additional examples of the mosaics that can be created with the FAMOS model:
\begin{itemize}
\item \href{https://photos.app.goo.gl/cyF4XvHWXU3bzwUW9}{Gallery 1: pure convolutional generation} (memory module disabled). When the style images have \textit{texture-like} properties, this architecture is also an interesting tool and can create good looking output image stylizations (texture mosaics), while being faster computationally. Please see Figure \ref{fig:la} for an example of the capabilities of that mode of FAMOS. 
\item \href{https://photos.app.goo.gl/rSJpy9ySTuyYiXV47}{Gallery 2: non-parametric mosaics with FAMOS} which make use of the memory module -- this allows to copy flexibly parts of the style images when necessary.
\end{itemize}

\begin{figure}[t]
    \centering
    \includegraphics[width=4.5cm]{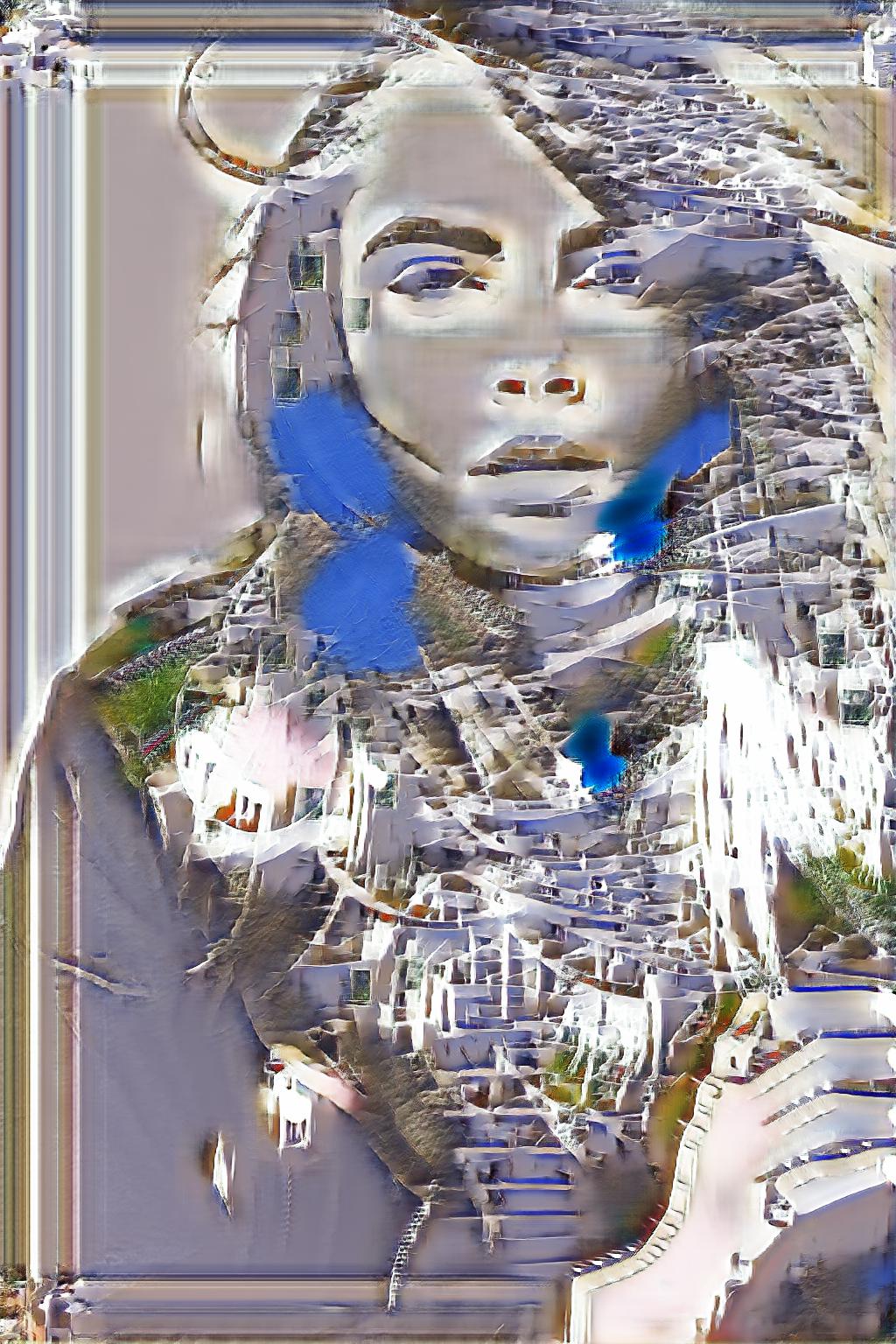}
    \includegraphics[width=4.5cm]{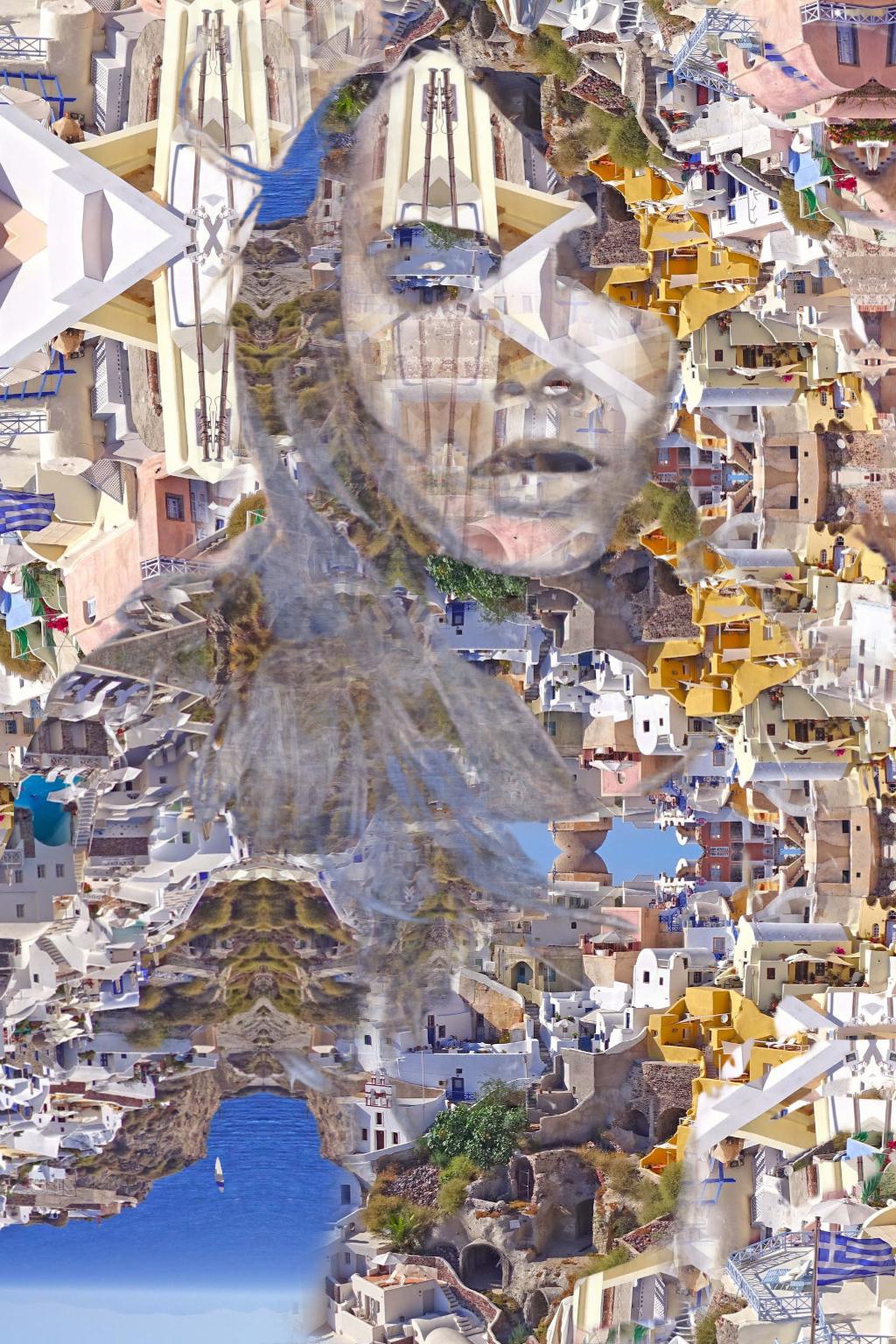}
    \includegraphics[width=4.5cm]{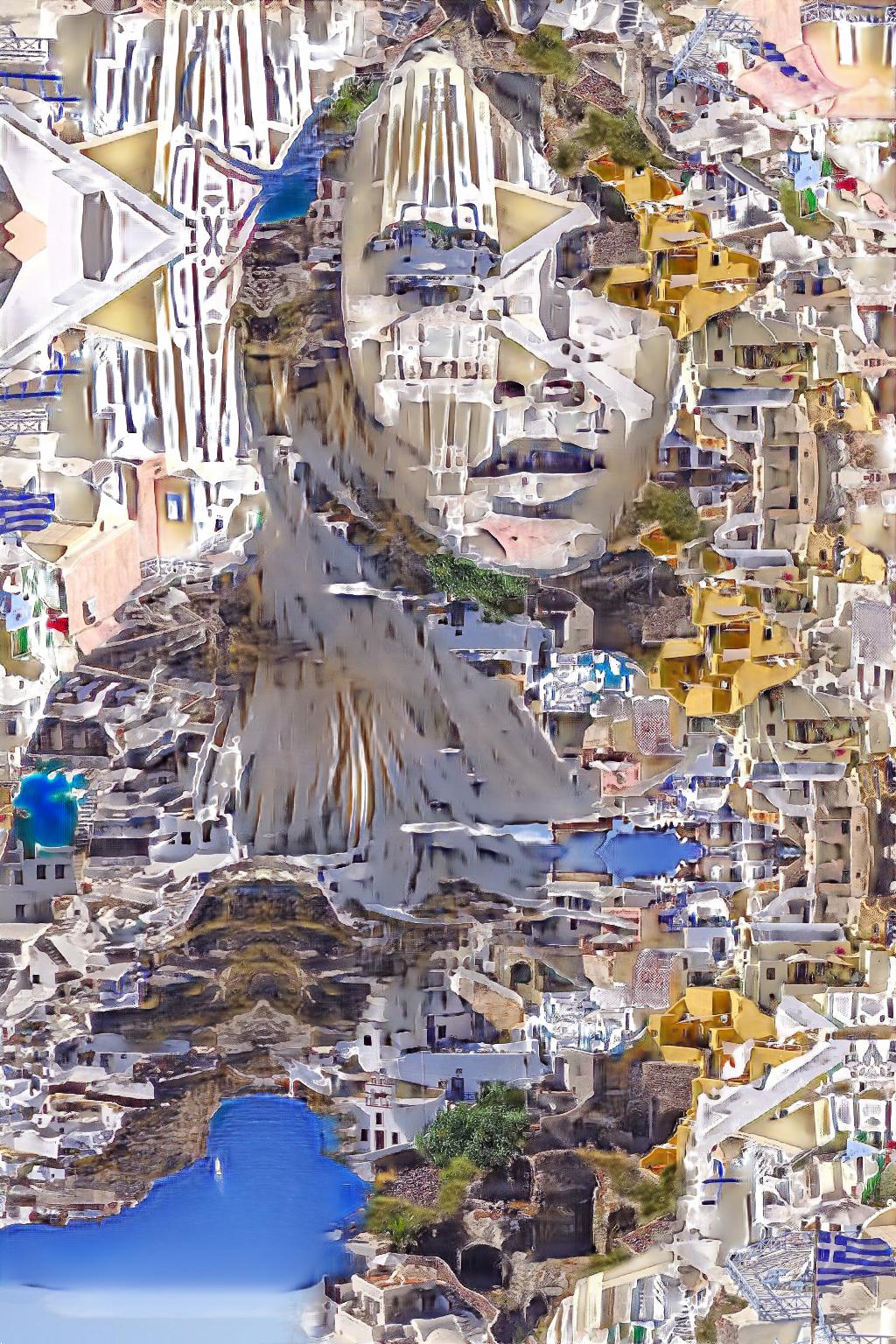}
    \caption{Different outputs from different variations of the architecture. From left to right: (i) only parametric image generation, (ii)  nonparametric mix without blending, (iii) blended parametric and non-parametric as in the full FAMOS architecture. Images of size 1536x1024 pixels, best seen zoomed-in.}
    \label{fig:ablation}
\end{figure}

Figure \ref{fig:ablation} shows the drastic differences in the output of the generative model from Figure \ref{fig:ablation} if we emphasize only some parts of it.
If we disable the memory $M$ and have $N=0$ templates (equivalently if $\alpha=1$ in the blending equation), we end up with a model very close to a traditional Unet generator, purely parametric generation of $I_G$. While in some cases this can be quite efficient as well, especially for easy to learn repetitive textures, it can fail for more complicated image styles such as the Santorini island one. 

We can also force $\alpha=1$ and keep entirely the template image $I_M$, a non-parametric behaviour. This can preserve well the details of Santorini, but has visual glitches: some hard edges, blurriness at borders of different template mixing regions.

The full output of FAMOS, the blended refined output has the best image quality in our opinion, keeping the details of the non-parametric mix and correcting some of the artifacts there. 

However, as might be expected of an artistic digital algorithm with so many options, depending on the complexity of the stylization image distribution, any of the 3 variants can be an effective tool of artistic expression.


\begin{figure}[t]
    \centering
    \includegraphics[height=4cm]{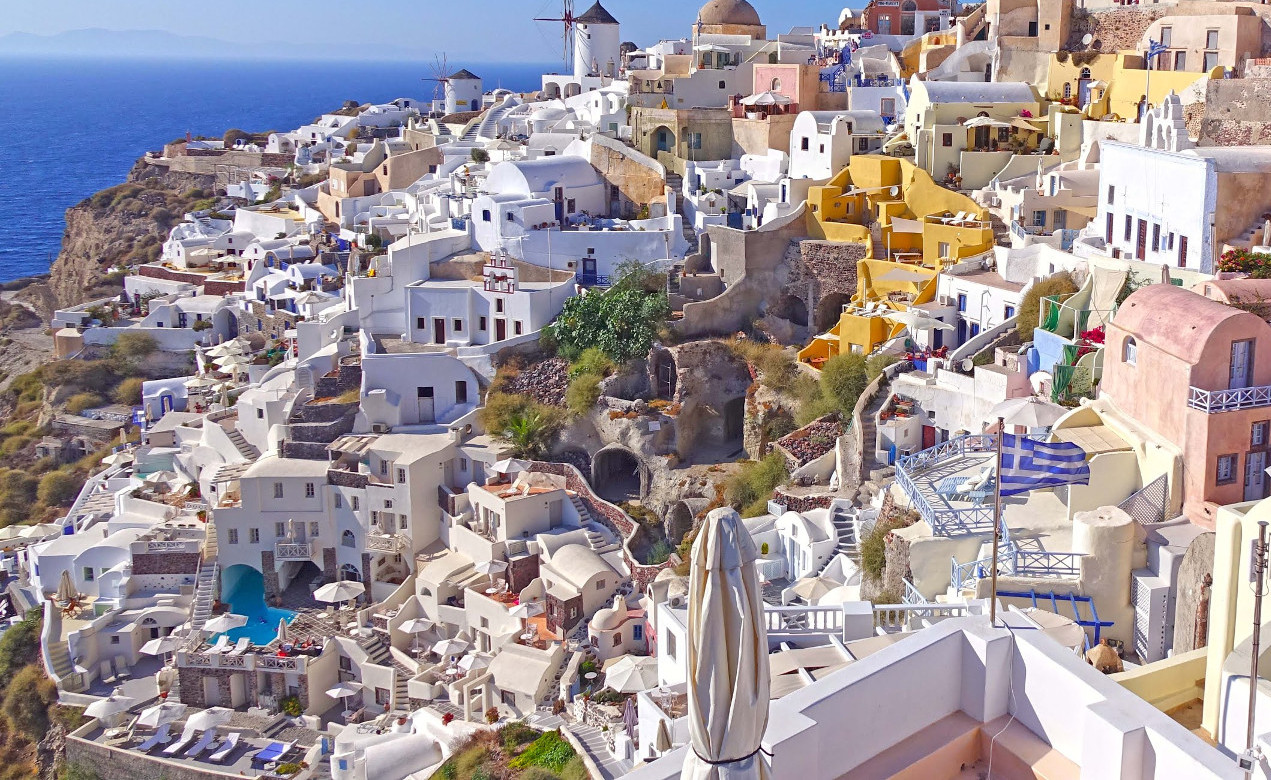}
    \includegraphics[height=4cm]{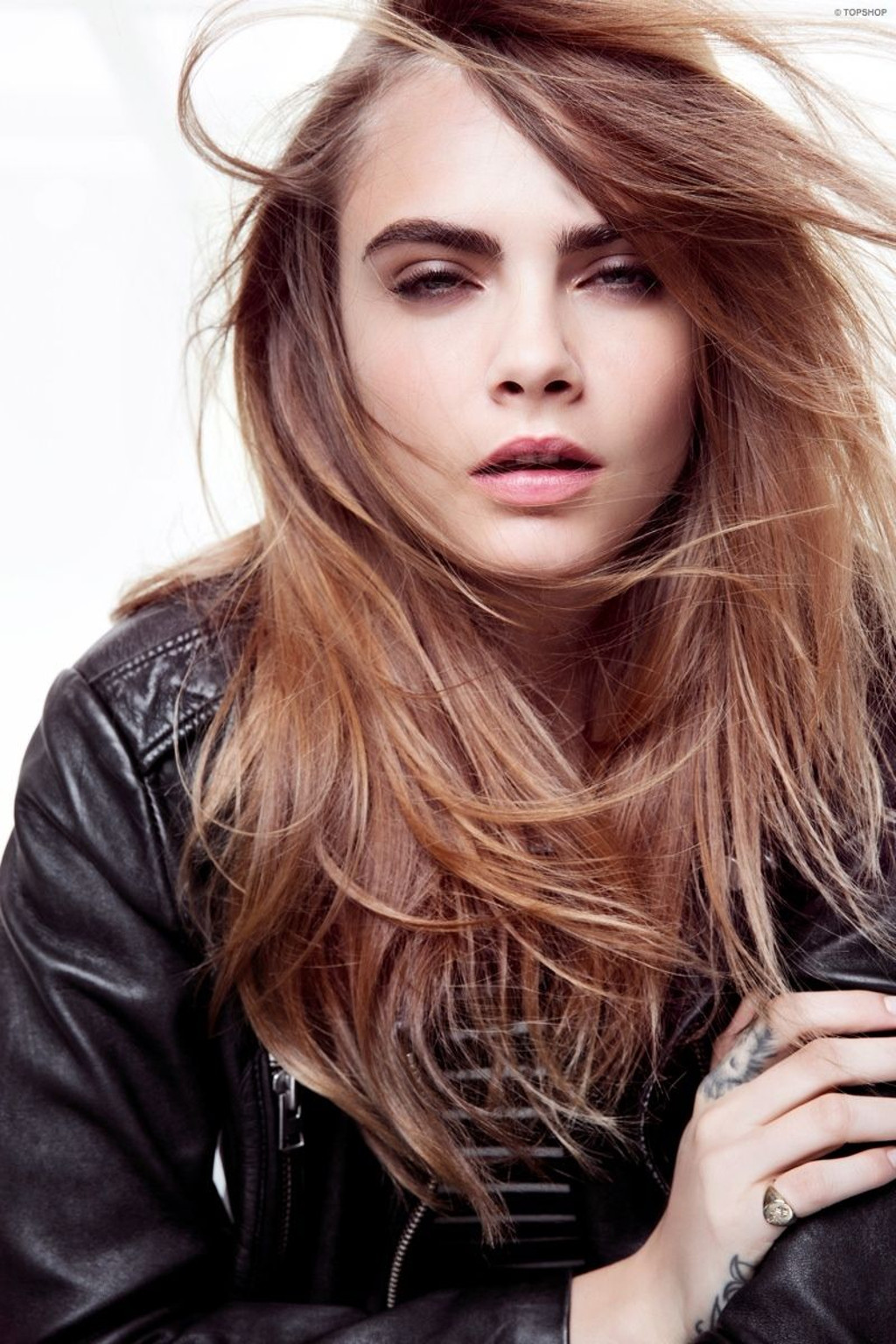}
    \caption{The image of the island of Santorini from \url{commons.wikimedia.org} we used as a complex stylization (unconventional texture since scale varies widly). The content image of a fashion model, from \url{www.zalando.de}.}
    \label{fig:inputs}
\end{figure}

\paragraph{Tiled memory templates}
The memory templates $M$ are created by moving a regular coordinate grid $\Theta = [-1,1  ] \times  [-1,1  ]$ from style image $I_T$.
 We can use either mirror or wrapping padding mode when interpolating using a coordinate grid $\eta+\Theta \in \mathbb{R}^{2\times H \times W} $, where $\eta \in \mathbb{R}^2$ is a random offset added to $\Theta$.
 
 In mirror mode (a.k.a. reflect mode), coordinate values $1+\epsilon$ are smoothly decreasing again and the interpolation routine would copy pixels from positions $1-\epsilon$.
 In mirror mode we are simplifying the task of the prediction module, since the templates tile neatly into each other without hard borders. This also has a very interesting visually kaleidoscope-like aesthetics because of the axes of reflection in the appearance of the tiled templates.
 
 In wrap mode, $1+\epsilon$ coordinates are hard reset and copy pixels from $-1+\epsilon$.
 In wrap mode, the model should learn and adapt to avoid borders, which will be penalized by the discriminator.
 
\begin{figure}[t]
    \centering
    \includegraphics[width=7cm]{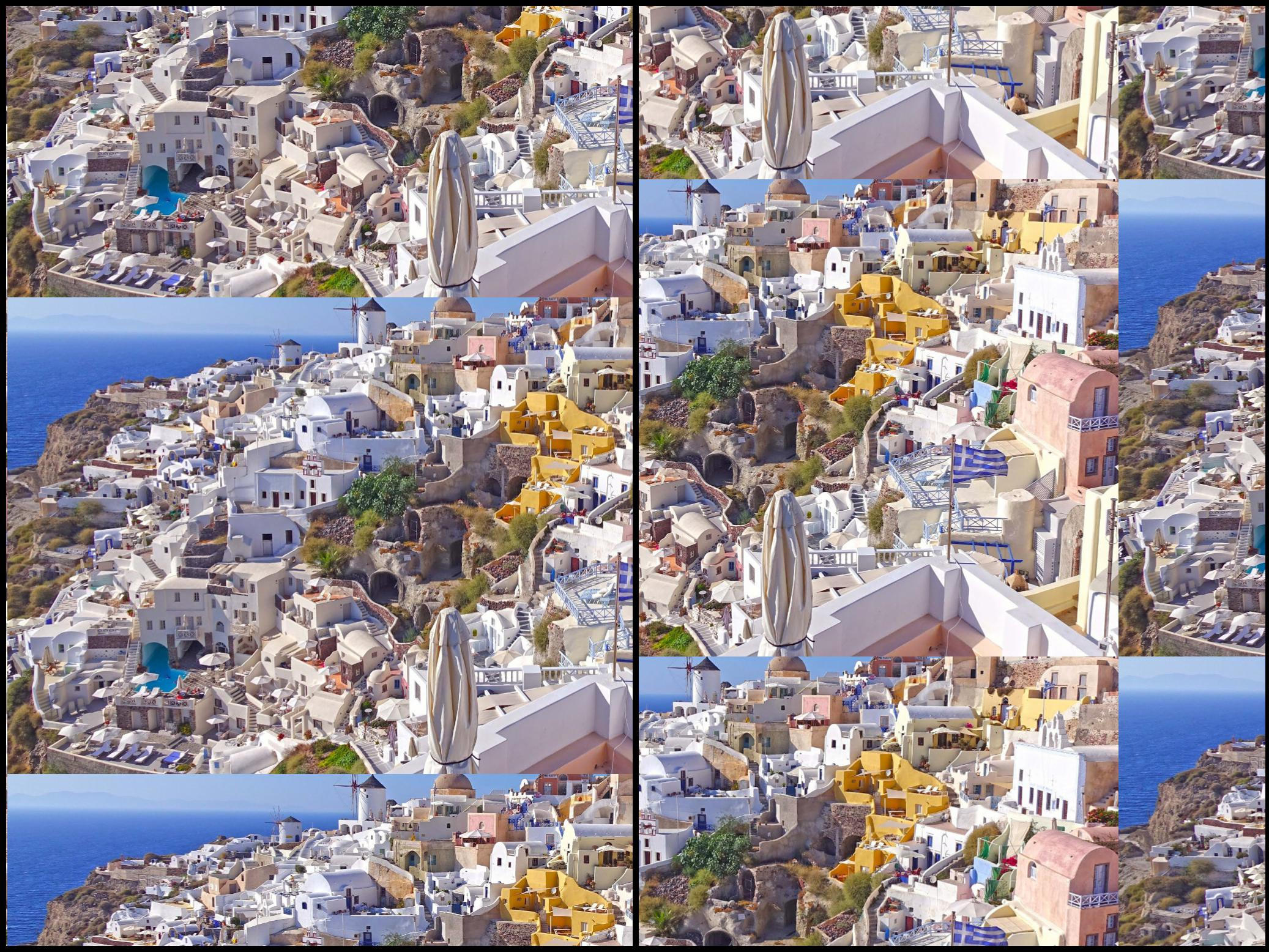}
    \vspace{0.5cm}
    \includegraphics[width=7cm]{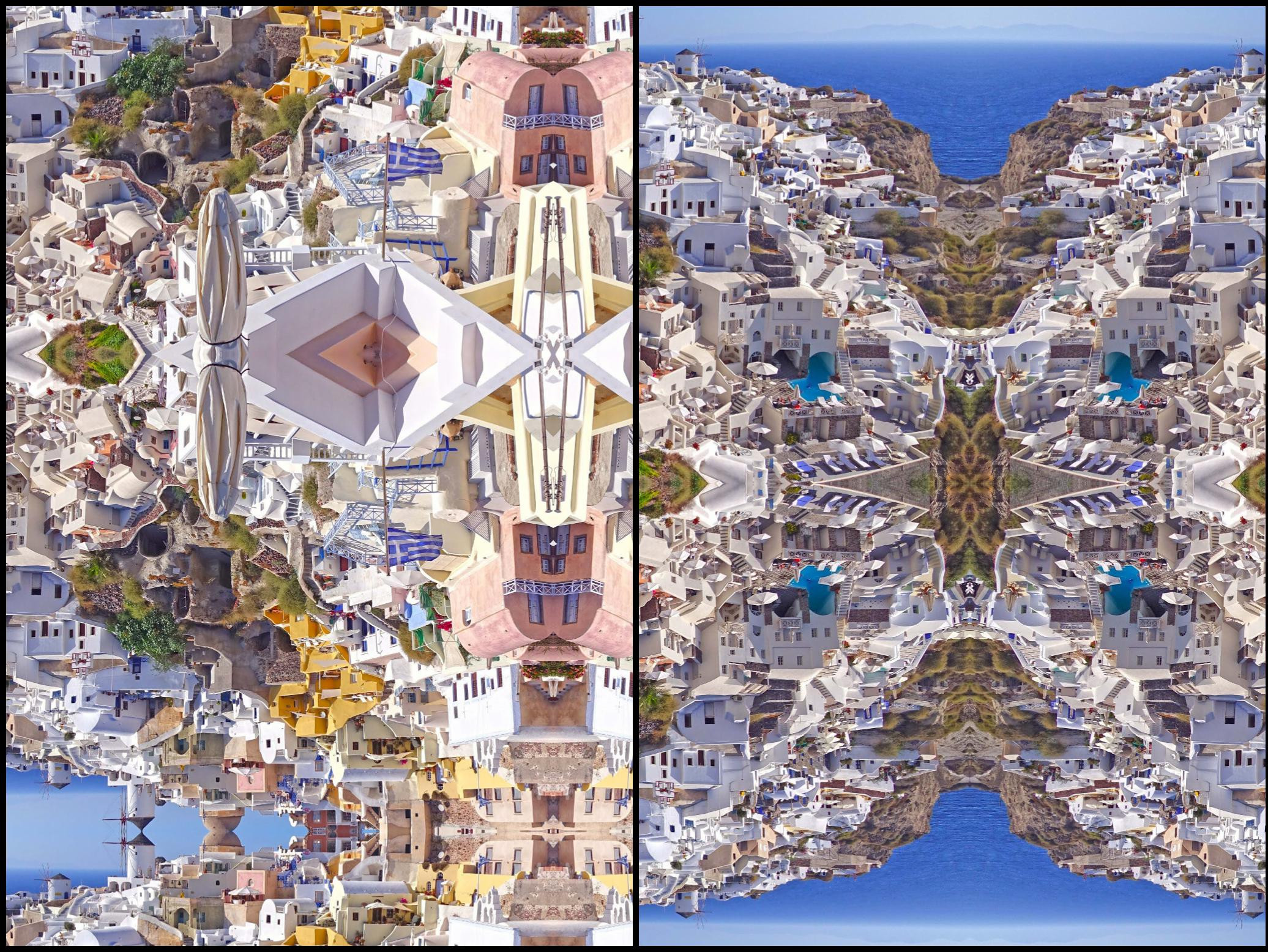}
    \caption{Using the Santorini panorama image as style (size 1270x780 pixels ) tilings of size 1024x1536 pixels (size of the given content image $I_C$) are  interpolated, 2 images for wrap and mirror mode illustration.}
    \label{fig:templates}
\end{figure}

 Figure \ref{fig:templates} illustrates this process, showing wrapped and mirrored mode, and 2 interpolations with random coordinate offsets to illustrate how this translation enriches the memory tensor $M$ and allows our fixed spatial position memory module to copy varied content.

\paragraph{Correspondence map}
The choice of \textit{correspondence map} used for reconstruction also matters, as discussed by \cite{GatysEB15a} who selects different layers of the pretrained VGG-19 network, or \cite{GANosaic} where the effects of dowmsamplng are discussed.
For FAMOS it worked well with any of these settings:
\begin{itemize}
\item convolve with a Gaussian filter (using reflection padding to avoid border artifacts) and convert to greyscale
\item downsample image 4,8 or 16  times and convert to greyscale
\item train a small \textit{reconstruction} conv. network with kernels 1x1 and stride 1 so that the loss becomes $\|I_C - \phi_{rec}(I)  \|$ -- effectively this converts from the color space of the textures to the content image.
\end{itemize}


\section*{Appendix II: Training and Implementation Details}
Our code would be released in GitHub at \url{https://github.com/zalandoresearch/famos} after publication at a conference or workshop.

\begin{figure}[h]
    \centering
    \includegraphics[width=7cm]{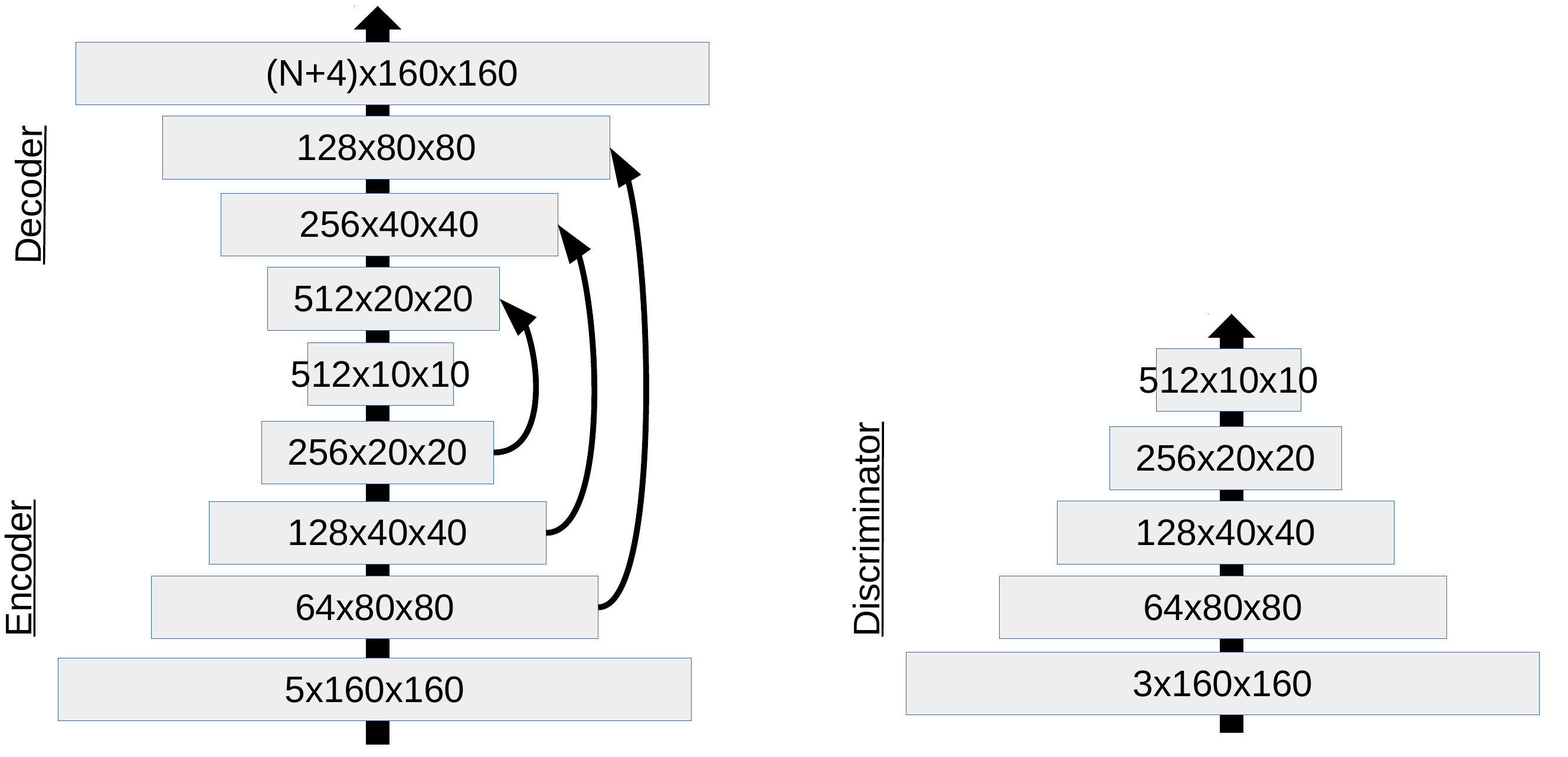}
    \caption{Channel and layer counts for our standard FAMOS architecture. We use typically 4 layer deep encoder and decoder (in Unet generator), and 4 layer deep discriminator. The generator takes content image $I_C$ and coordinates $\psi$ as input, and outputs the mixture array $A$, the blending mask $\alpha$ and image $I_G$ to use for blending. The total number of output channels is $N+4$, given $N$ memory templates.}
    \label{fig:channels}
\end{figure}

\paragraph{Network and training}
We used typically $N=100$ memory templates for $M$. Note that these can take a lot of memory if we have large spatial size $H,W$, but they can stay in RAM -- only patches from them are shifted to the more limited GPU memory.
Both generator and discriminator use batch norm and kernels of size 5x5. We use ReLU and leaky ReLU nonlinearities. To avoid checkerboard artefacts we use upsampling-convolution, instead of transpose convolution.
The training patch size was 160 -- but deeper Unets would required larger patch size; the minibatch size was 8.
The typical channel (width )and layer (depth) counts we used for generator and discriminator are shown in Figure \ref{fig:channels}.
Standards architectures as described in \cite{pix2pix} can also work.
For training we use DCGAN loss \cite{RadfordMC15} and the ADAM \cite{KingmaB14} optimizer.

Our code is implemented in Pytorch, and we ran it on a single NVIDIA p100 gpu card.
The time to get first nice image is a few minutes usually, but several hours can be required to train fully a complicated model. 



\paragraph{Note of caution: training instabilities}
 We also note in general that the FAMOS architecture can sometimes diverge or require some tuning given different  content and style images. The mixing properties of the texture process of the style image are very important (see for some discussion \cite{PSGAN2017}). Such texture properties determine how easy it is to sample and generate -- a simple texture (e.g. rice particles) is much easier to use than a non-texture image such as the island of Santorini from Figure \ref{fig:inputs}. The interplay of generation and curation on the human artist side is an essential part of generative art, as a human counterpart to the adversarial nature of the GAN game between generator and discriminator. Changing the parameters or restarting can often lead to better results. As with many GAN models, early stopping may be useful: the user of FAMOS can save regularly the output of the model and keep those images that seem most promising.
 
A particular failure mode can happen if the mixing matrix $A$ collapses: few templates from $M$ are always used to generate $I_M$ and the other templates are ignored (i.e. entropy of $A$ close to 0). In that case, the parametric part of FAMOS can still paint a nice image on top of that $I_M$ serving as background canvass, but it would be preferable if FAMOS can use the full expressive power of it memory templates. The other extreme case happens rarely: the entropy of $A$ stays high and all templates mix to a greyish image $I_M$. This may require special tuning of an entropy regularization schedule. 

We acknowledge these convergence issues, but even so we think that FAMOS is an interesting novel image generation method, and is a fun tool to use and explore. Some regularization terms can help stabilize training of FAMOS, but the exact research of the "right" regularization terms and weight in the loss is left to future work. We considered these terms as part of a regularization loss $\mathcal{L}^{regularize}$ for the generator.

\begin{itemize}
\item small entropy of mixture matrix $A$ -- this will force values to be 0 or 1, allowing to cleanly select and keep some memory template, avoid blurriness
\item small total variation of $A$ -- smooth changes
\item small norm of blending mask $\alpha$ -- i.e. forcing that we are close to $I_M$, the true memory templates, and paint less with the parametric GAN. Thus will force to only paint details with the GAN Unet when copying from memory does not work
\item small total variation of $\alpha$
\end{itemize}

While it is not entirely clear when these terms stabilize training, they are interesting on their own as additional controls from side of the artist user of FAMOS.
The next paragraph has more comments on the stability and performance of the method.

\begin{figure}[t]
    \centering
    \includegraphics[width=8cm]{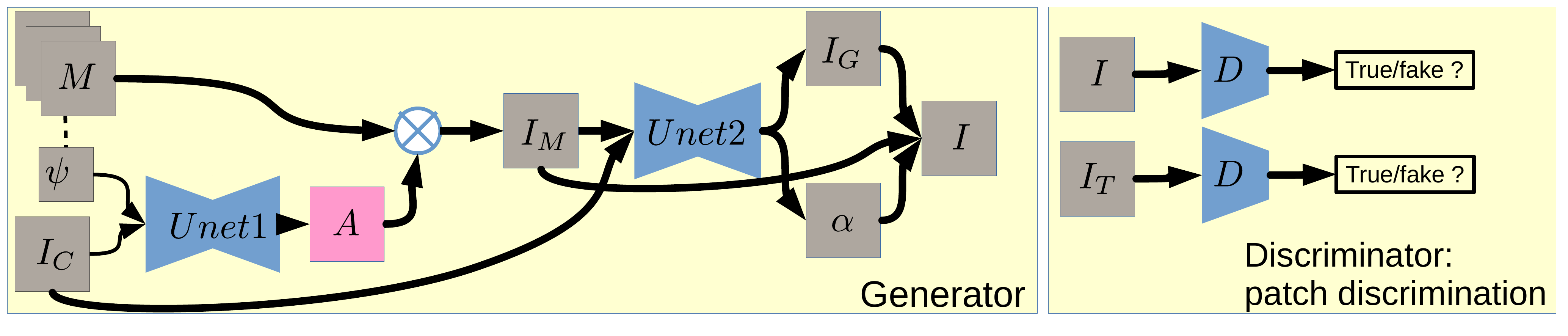}
    \caption{An alternative architecture of the generator $G$ with a chain of non-parametric and parametric modules. The discriminator of true/fake looking texture patches is standard patch-based one \cite{pix2pix}. }
    \label{fig:archold}
\end{figure}

\paragraph{Practical tips: what works and does not work}
In order to find a good architecture for FAMOS, we tested a lot of architectural choices. Some work better than others.
\begin{itemize}
\item If the cropping coordinates $\psi$ are identical for content and template image patches distributions $P_C,P_M$ -- generalization to out-of-sample content images may deteriorate since the network learns by heart that some content patches go together with some template patches. To remedy this, cropping content patches and memory templates at different random locations allows better generalization.

\item Downsampling and copying the coordinates $\psi$ after every pair of  conv+batch-norm layers makes FAMOS better when using this croppng mode for generalization. This effect is subtle and needs more investigation, but we assume that it makes the network more sensitive to the cropped template offset location.

\item However, if we want to train really well just mosaics for the training  content image set, and do not need to generalize to additional out-of-sample content images, then we can crop the "same" coordinates from $M$ and $I_C$, both for training and inference, thus allowing the network to learn much better the spatial relation between content and memory template, and give better mosaics result. In a sense, such a mode is analogous to optimization based stylization \cite{GatysEB15a,GANosaic}, since only the result on a single image matters. However, even in that case FAMOS remains a very performant model capable of dealing with very high resolution content images.

\item We use a single Unet to predict the $N$ channels of $A$ for mixing templates and the 4 channels of $\alpha,I_G$ for blending -- this additional weight sharing works and is much faster than having a design with 2 Unets as in Figure \ref{fig:archold}. But further experiments may find cases when more capacity (e.g. by adding residual layers) can improve image quality, as is often the case for GAN methods. 

\item WGAN-GP \cite{wgan} loss gives better convergence to FAMOS than DCGAN loss \cite{RadfordMC15} when having nonparametric memory $M$. However, if we disable the memory and rely only on the parametric part (e.g as in Figure \ref{fig:la}) than DCGAN is much better. This inconsisteny which loss is best was quite surprising for us, but we did not investigate it in detail.

\item Adding some noise channels (identical spatially and of spatial dimension as $I_C$) to the input of the Unet (concatenated to $I_C,\psi$) or to the bottleneck, seems to stabilize the behaviour of the Unet. However, the outputs are not very sensitive to the stochastic noise inputs -- some techniques \cite{modal} exist to ensure that conditional GAN methods have multimodal outputs. 

\item If $M$ is an input to the generator as well, $3N$ more channels when having $N$ templates -- no gain in quality, only downside of more computation required. 

\item We use randomly translated copies of the textures in $M$ and duplicate the images with different offset. An alternative can be to predict the morphing to each of the texture images and avoid having random duplicates. We tried directly morphing with optical flow -- there were issues with poor gradients of the loss and deformations of the style images, so we did not pursue that option.

\item If use only a supervised content loss without GAN loss for the memory mixing module -- poor results, the GAN loss is required indeed.

\end{itemize}

\section*{Appendix III: Outlook}
As stated in the previous section, we are examining in detail various model convergence issues and testing regularization terms that can improve stability of FAMOS.

In the future, we plan to investigate the ability of our model to generalize to new content and memory images at inference time. The capacity of our model (channels of the Unet) with respect to the template memory size $N$ can be also examined.

We currently mix the memory templates only at a single scale using RGB pixels. Previous work \cite{liwand16} has used multiple scales and feature spaces other than RGB for copying -- this can be added as improvement in the FAMOS framework.

The U-net we use for mixing coefficients prediction can be replaced with an attention-like structure \cite{NonLocal2018}, which can further improve the generalization of the FAMOS model, allowing to use at inference time many additional texture templates. Our first results seem promising in that direction.



\end{document}